# A mixed-methods ethnographic approach to participatory budgeting in Scotland

JONATHAN DAVIES

Computer Science, Warwick University, Coventry, UK, jonathan.davies.1@warwick.ac.uk

MIGUEL ARANA-CATANIA

Computer Science, Warwick University, Coventry, UK, miguel.arana-catania@warwick.ac.uk

ROB PROCTER

Computer Science, Warwick University, Coventry, UK & Alan Turing Institute for Data Science and AI, rob.procter@warwick.ac.uk

FELIX-ANSELM VAN LIER

Government Outcomes Lab, Oxford University, UK, felix-anselm.vanlier@bsg.ox.ac.uk

YULAN HE

Computer Science, Warwick University, Coventry, UK, yulan.he@warwick.ac.uk

**ABSTRACT**

Participatory budgeting (PB) is already well established in Scotland in the form of community led grant-making yet has recently transformed from a grass-roots activity to a mainstream process or embedded 'policy instrument'. An integral part of this turn is the use of the Consul digital platform as the primary means of citizen participation. Using a mixed method approach, this ongoing research paper explores how each of the 32 local authorities that make up Scotland utilise the Consul platform to engage their citizens in the PB process and how they then make sense of citizens' contributions. In particular, we focus on whether natural language processing (NLP) tools can facilitate both citizen engagement, and the processes by which citizens' contributions are analysed and translated into policies.

## 1 INTRODUCTION

In 2017 the Scottish Government and the Convention of Scottish Local Authorities (COSLA) agreed that at least 1% of the total local authority budget (at least £100 million) will be decided by communities through participatory budgeting (PB). This landmark agreement marks a shift from what was once a highly localised, community led process to one situated and embedded in mainstream Scottish policy and politics. PB has become a policy 'instrument' used to

"improve decision making and tackle inequalities", rather than a symbolic or technical 'device' for community engagement [14 p. 11]. In this form, this instrument is a new way of governance which requires a reconfiguration of the relationship between local authorities, society and the data produced.

This ongoing research paper analyses the implementation of a digital participation platform and natural language processing (NLP) technology to allow for the effective use of PB as an instrument. First, we provide a brief introduction to PB in general and to PB in Scotland, following the work of Harkin et al. [14], Escobar et al. [9] and O'Hagan et al. [20]. We then explore digital PB, specifically the role of the Consul digital participation process that is being used in the mainstreaming process. For the research methodology, we employ a mixed-methods, ethnographic study of the 32 local authorities to determine how each one utilises the Consul platform, makes sense of citizens' proposals, translates them into policy and measures the impact of the process. This method allows for a deeper understanding of the 'local knowledge' [12] generated from the PB process and a richer evaluation beyond surface level observations (for example, looking at proposals made alone). If this initial, small scale PB exercise is assessed as being successful, the Consul platform will become the vehicle for further roll out of PB processes in Scotland on decisions on over £10 billion of public sector spending by 2022.

## 2 PARTICIPATORY BUDGETING IN SCOTLAND

PB first emerged in the Brazilian city of Porto Alegre in 1989 as a way to tackle democratic deficits and better support marginalised communities [1, 21]. This local process soon grew to a global movement with an estimated 11,690-11,825 PB cases worldwide today [7]. This rapid global diffusion has made it difficult to agree upon a concept or single definition which may be applied to all budgetary decision-making processes. Instead, we highlight three key characteristics that are present across all PB initiatives:

- Many participatory processes are designed to engage with certain expert groups, stakeholders, or a cross-section of the public (for example focus groups, consultations, or citizen assemblies). PB instead engages directly with citizens, placing them at the centre of the process. As Smith [22] points out, experts or stakeholders are also members of the public. However, PB is innovative in the fact it engages with citizens because they are citizens. Escobar [8] builds upon this notion to use the term citizen in an expansive sense. Citizenship in PB transcends the common or legalistic definition to include typically excluded groups such as refugees or children.
- PB involves a commitment to a specific portion, or the entire amount of an institution's budget. This separates PB from many other participatory processes which are designed to collate ideas with a promise of action dependent on top-level agreement.
- Citizens are participating directly in making decisions. Yet key actors of the representative system also play a central role in this direct democratic process. Viewing the process through this lens presents it as something more than a reallocation of budgets. It becomes a collective process of shared power and creates new connections between citizens, political representatives, and local government officials. As Escobar [8] illustrates, the process is a form of co-production.

To understand PB in Scotland we must first explore the various political, institutional, and social factors which underpin its emergence. Firstly, political events such as the Scottish independence referendum in 2014, the EU 'Brexit' referendum in 2016 and the recent UK General Elections have evoked issues of national identity and dissatisfaction with democratic structures [16]. These concerns are, secondly, complemented by pre-existing institutional factors which result in Scotland having "the largest average population per basic unit of local government of any developed country" [9 p 312]. Just 32 councils serve a population of 5.4 million, resulting in an average of 168,750 citizens per local authority (compared with the EU average of 5,615 citizens) and a ratio of 1 elected councillor each representing 4,270 Scottish citizens (compared with a ratio of 1:700 in Spain, 1:500 in Germany, and 1:400 in Finland) [9]. Finally, despite clear potential for a greater disconnect between communities and local authorities there is a strong social desire for greater participation in local decision making (see [16] and [17] for two recent surveys on participation) and civic engagement [13]. Together, these factors have generated a turn towards public service reform, community empowerment, and desire for democratic renewal which PB has been at the heart of [9].

The first wave of PB processes in Scotland is commonly referred to as 'first generation PB' [14]. This has been characterised by community grant-making emerging organically from grass-root organisations, when there was suitable support and community priorities matched available funding [9]. This grassroot momentum soon gained increasing political, legislative and policy support [6]. As part of this turn, the Scottish Government, through the Community Choices Fund, supported a national development programme to underpin the introduction of PB in partnership with local authorities [20]. This represented the embedding of PB across Scotland and a move to what is commonly described as 'second generation PB' or mainstreaming. This was further strengthened in 2017 when the Scottish Government and COSLA agreed that at least 1% of local authority budget will be decided by communities through participatory budgeting. This agreement marks an important juncture linking 'second generation PB' back to the characteristics explored above. Firstly, mainstreaming PB places all citizens at the centre of the process, beyond grass-root organisations. Secondly, it is a commitment by local authorities for a set amount. Thirdly, it is not simply an increase in resource availability but instead represents a space for co-production due to new connections made between citizens and local authority officials.

## 3 DIGITAL PARTICIPATORY BUDGETING AND NLP

The rapid global uptake or scaling up of PB is gradually leading a number of governments to use digital platforms as a primary means of participation in this process. "My Neighbourhood", the Icelandic portal using the free software technology of Better Reykjavík, was one of the first successful use cases of digital platforms for participatory budgeting as early as 2011. This case served as a direct inspiration for the development of the free software platform Consul, initially used in the city of Madrid in one of the largest participatory budgeting cases worldwide with 100 million euros per year. This platform was subsequently extended to dozens of administrations around the world and, in particular, closing the circle back to the birthplace of participatory budgeting in Porto Alegre.

While the use of digital platforms has only begun to take off in recent years, in some cases of broad participation, effects related to this can already be observed. Other direct democracy mechanisms such as Citizens' Initiatives or Referendums focus the large-scale participatory process on the stages of supporting or voting on proposals, which can easily be aggregated regardless of the scale of participation. In the case of participatory budgeting, although this depends on each specific participatory design, in general each participant can issue their own proposals, making the complexity of the process grow in direct proportion to the number of participants. In the case of offline processes, this has been usually easily solved due to the low number of participants by facilitating face-to-face deliberative spaces. However, once the barriers to inclusiveness of face-to-face participation are removed, digital participation results in an exponential growth in the number of participants. In the case of Madrid, for example, more than 90,000 participants registered in 2018. These levels of participation allow us to begin to observe the effect of information overload on the process [2].

To tackle the problem of information overload in participatory budgeting from its earliest appearances, we developed a range of NLP-based tools and integrated them into the Consul platform [see 2]. Evaluation of these tools has produced encouraging results, but they have not yet been tested in real world scenarios. The Council of Scottish Local Authorities (COSLA) approached us about deploying our tools in a rolling programme of participatory budgeting (PB) that COSLA is launching in 2021. This represents an opportunity to beta (i.e., live) test these NLP tools and discover if they have value for improving citizen engagement in PB, while also identifying factors that will ensure that local authorities in Scotland and the rest of the UK are able to scale up PB in the future.

The NLP tools have been designed to tackle stages in the process of participation and interaction with the Consul platform where the amount of information available is difficult for users to process, or where users find it difficult to interact with each other. In particular, the following processes were identified as being highly complex in terms of information: reading proposals and identifying global categories in which to classify them, identifying each proposal in one or more of these categories, detecting similarities between proposals allowing all similar proposals to be presented together, summarising the comments made on each proposal, identifying users with similar interests to facilitate their debate and collaboration [2].

To facilitate these processes, the following NLP techniques were implemented on the Consul platform: topic modelling of the proposals, text summarisation on the comments, topic modelling of the user generated content.

These developments therefore have the potential to facilitate more effective and intelligent interaction both between users and with the content created in the deliberative process. The goal is not to extract intelligence from users, but to make the processes themselves smarter.

However, if solving the information overload problem experienced by citizens leads to more active engagement and therefore more ideas being generated, will the result be simply to transfer the information overload problem to those officials responsible for the sense making and translation of into policy? Some COSLA member councils have been reluctant to trial Consul or some of its functionality (e.g., Debates and Proposals) as local officers do not have capacity to analyse and summarise large amounts of data 'by hand'. There may therefore also be a need for NLP tools to provide assistance in the analysis of the opinions and ideas generated by citizens.

## 4   RESEARCH METHODOLOGY

This project draws on a mixed-methods ethnographic approach to move beyond a thin, or surface level, observation and instead capture the deeper or thicker contexts to which local authorities make sense of new forms of local knowledge generated through the Consul platform.

Each of the 32 local authorities will individually implement and run a process of PB through the Consul platform, with some using the standard version and others the version enhanced with the NLP tools. This provides the unique opportunity to investigate the PB process and its outcomes in 32 sites, each with distinct economic makeup, demographics, administrative structures and practices, and thus the impact of such factors on the policy outcomes. This comparative contextualisation closely follows a Geertzian approach providing 'thick comparisons' of different processes [15]. As Geertz states, comparison of one site should not be made in relation to a generalized form but instead relative to another locality [12]. Doing so highlights phenomena which may otherwise be missed.

A mixed-methods design enables us to identify and measure the impact that the NLP tools have on the process and on its outcomes, including any deficiencies that would point to the need for further refinement and development. In-depth, secondary research will first be carried out to provide contextual information on the current state of PB in Scotland. We will then observe local authorities through the attendance of meetings related to the management of the PB process and semi-structured interviews with key actors to understand how sense is made of this new source of local knowledge. A thematic analysis [11] will be employed to identify and report on common themes that emerge. Finally, quantitative assessments of platform content, such as the type of proposals made, will be informed by the qualitative rich data and vice-versa, filling potential knowledge gaps and providing a deeper and richer analysis.

## 5 POTENTIAL IMPACT

In its first generation, PB remains a peripheral device for highly localised, smaller grants. To become an embedded policy instrument requires a reconfiguration of the relationship between local authorities, society and the data produced. This involves greater collaboration and commitment between these actors which can then open up space for generating local knowledge (including a reconsideration of citizen priorities to better tackle inequalities and an increase in transparency and understanding of local authority spending).

Secondary research has so far shown that there is an appetite for greater collaboration and a strong desire from local authorities for the mainstreaming of PB in Scotland. Initial interviews with council officials have confirmed this enthusiasm yet have highlighted a number of key challenges which have arisen from community led grant making processes, or first generation PB, when run online. This includes a strain on the local authority workforce (due to, for example, the manual filtering of proposals or the voting process), a resistance from elected council members, and a lack of deliberation or discussion during the process itself.

The challenges experienced are akin to ones observed across other studies of online participatory processes. This includes a lack of skill or resources to process citizen suggestions effectively [5], difficulties in coordinating political action and collaboration [10], and information overload in online deliberation [21]. Previous analysis of the Consul platform, which used NLP in a controlled use case, shows an improvement in the participatory process through affording participants improved information analysis and interaction [2]. We therefore argue that the use of the Consul platform, as an interface between active networks of communities, citizens, and local authorities, can impede these common challenges by providing the necessary space for collaboration and co-production.

The next stage of the investigation will be to develop a deeper understanding of how each local authority will use the Consul platform to combat the challenges raised, allow for the generation of local knowledge and make sense of it. This will allow COSLA to provide better PB service support to its member councils and allow these councils to provide a better service to their citizens and communities. If shown to be successful, the enhanced Consul platform will become the vehicle for further roll out of PB processes in Scotland on decisions on over £10bn of public sector spending by 2022. The enhanced platform will also then be made available for Consul's existing user base (135 institutions in 35 countries) and future users to download and install.

To support this scaling up of PB, we will aim to develop a broader framework on how to meaningfully collect and analyse large-scale public data in participatory processes of similar complexity and scale, and how to employ tools such as Consul effectively at different stages of the process. This will also highlight the limits of such tools and propose methods and strategies for assessing the impact of digital tools on such processes.

Beyond addressing issues of process design, our study will foster more nuanced knowledge among data scientists when it comes to the design of effective and legitimate participation and governance processes, more broadly; and among politicians, lawyers and public sector workers about the opportunities and limits of digital participation tools. This will trigger broader theoretical improvements concerning procedural fairness, openness and potential biases in tech-enabled policy-making, and the changing role of representative democracy in the digital era.

**ACKNOWLEDGMENTS**

This research has been supported by the Alan Turing Institute for Data Science and AI (grant no. EP/N510129/1) and by the EPSRC Impact Acceleration Fund (grant no. G.CSAA.0705).